\documentclass[10pt,letterpaper]{article}

\usepackage{cogsci}

\usepackage{graphicx}
\usepackage{placeins}
\usepackage{enumitem}
\usepackage{amsmath}
\usepackage{xcolor}

\cogscifinalcopy 

\usepackage{pslatex}
\usepackage{apacite}
\usepackage{float} 

\DeclareMathOperator*{\argmax}{arg\,max}

\title{Show or Tell? Demonstration is More Robust to \\ Changes in Shared Perception than Explanation}
 
\author{

{\large \bf Theodore R. Sumers (sumers@princeton.edu)} \\
  Department of Computer Science, Princeton University \\
  Princeton, NJ 08540 USA

\AND 
{\large \bf Mark K. Ho (mho@princeton.edu)} \\ 
{\large \bf Thomas L. Griffiths (tomg@princeton.edu)} \\
  Department of Psychology, Princeton University \\
  Princeton, NJ 08540 USA

}

\begin{document}

\maketitle

\begin{abstract}
Successful teaching entails a complex interaction between a teacher and a learner. The teacher must select and convey information based on what they think the learner perceives and believes. Teaching always involves misaligned beliefs, but studies of pedagogy often focus on situations where teachers and learners share perceptions. Nonetheless, a teacher and learner may not always experience or attend to the same aspects of the environment. Here, we study how misaligned perceptions influence communication. We hypothesize that the efficacy of different forms of communication depends on the shared perceptual state between teacher and learner. We develop a cooperative teaching game to test whether concrete mediums (demonstrations, or ``showing'') are more robust than abstract ones (language, or ``telling'') when the teacher and learner are not perceptually aligned. We find evidence that (1) language-based teaching is more affected by perceptual misalignment, but (2) demonstration-based teaching is less likely to convey nuanced information. We discuss implications for human pedagogy and machine learning.

\end{abstract}

\section{Introduction}
Humans leverage a vast body of shared experience and knowledge when conducting and interpreting communicative acts~\cite{tomasello}. But communicative acts come in different forms, each with their own strengths and weaknesses. Language, particularly in the form of explanation, can be a powerful and efficient means of conveying abstract information about categories, relationships, and causal structures in the world. However, interpretation of language relies heavily on shared context between a speaker and listener~\cite{grice1975logic, ClaWilk, clark1996using, goodman_2016}. Without context to ground the meaning behind linguistic acts, language can break down and become ineffective. As an example,  consider trying to explain Einstein's Theory of General Relativity to someone with minimal mathematical or physics background. 

At the same time, people readily use non-verbal means of communication such as gesture~\cite{goldin_meadow_1999} or demonstration~\cite{gergely_2006} in adaptive, context-specific ways to convey relevant information about the world. Non-verbal communication lacks the combination of expressive richness and precision of language (imagine conveying the Theory of General Relativity by acting it out instead), but it also does not require the same foundation of shared experience. This may be why even 14-month-old infants show a capacity to reason about the complex interaction of intentions and physical context to recover the meaning behind communicative demonstrations~\cite{kiraly_2013}. 

Recent work on human pedagogy has examined the role of  language~\cite{Chopra2019}, demonstration~\cite{ho_2016}, and asymmetric information~\cite{velez_2018}. However, comparatively little work has studied the interplay of these factors. Here, we investigate the interaction of shared context with different forms of communication, exploring their efficacy and robustness to changes in shared perception.  

To address these issues, we develop a new cooperative teaching game. We provide teachers and learners with different user interfaces and allow communication via chat messages or demonstrative play. We test differences in shared context; in particular, we induce perceptual misalignment by ablating visual information from the task interface. We show that linguistic teaching suffers when teachers have limited access to the learner's perceptual state, whereas demonstration does not. Intriguingly, we also find evidence that language facilitates transmission of more nuanced concepts.

\begin{figure*}
\begin{center}
\includegraphics[width=18cm]{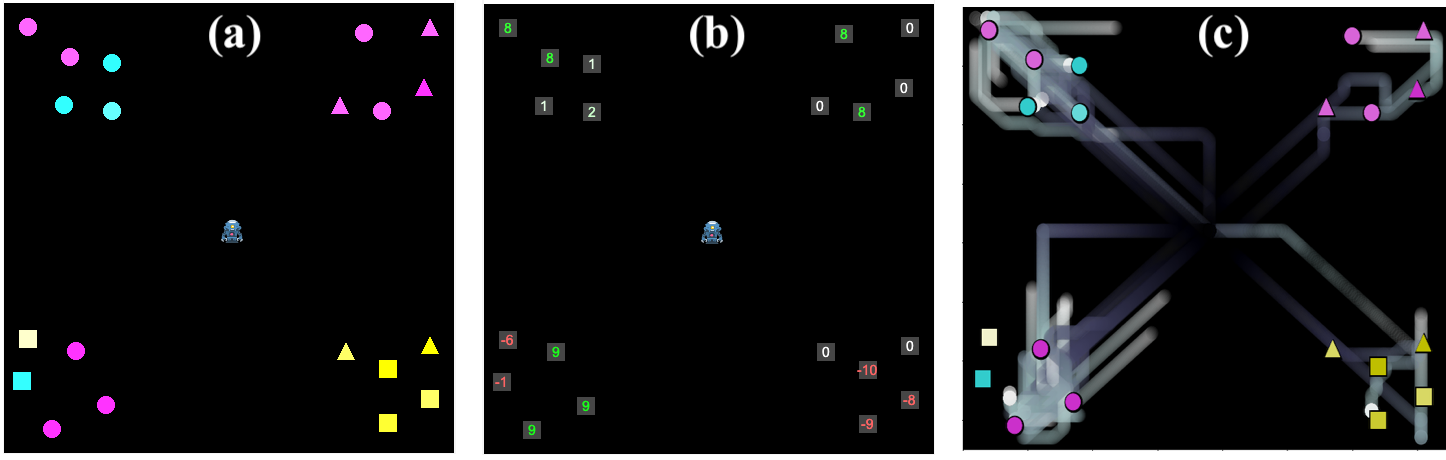}
\caption{Our collaborative teaching game, which gave players 8 seconds to move a robot and collect objects. (a) All learners saw this display, which masked object values with the target concept shown in Figure~\ref{value_map}. (b) All teachers saw this display, which showed the underlying object values. ``Full'' visibility teachers saw both (a) and (b) side-by-side and thus shared percepts with the learner. ``Partial'' visibility teachers saw only (b), and so did not share color and shape percepts. In the single player conditions, ``Solo-Partial'' players saw only (a), while ``Solo-Full'' saw only (b). (c): 25 randomly sampled player trajectories. Each reflects a particular belief about the target concept. Intuitively, choosing the top-left cluster indicates a belief that circles are valuable; the top-right that pink objects are valuable, the bottom-right that yellow objects are valuable, and the bottom-left that pink circles are the most valuable (note all players here choosing the bottom-left avoided the negative objects there).}
\label{game_ui}
\end{center}
\end{figure*}

\section{Collaborative Teaching Experiment}
Teaching games can be used to model concept transmission between two parties \cite{shafto_2014}. Subsequent research has explored sequential teaching \cite{macGlashan_2017, rafferty_2016}. This work generally uses a single communication mechanism and does not study misaligned perception. 

We develop a sequential teaching game to study (1) variable visibility into the learner's perceptual state and (2) different communication mechanisms. Basic gameplay, shown in Figure \ref{game_ui}(a), consists of one participant (the learner) moving a robot character to collect objects of various shapes and colors. Each object is worth -10 to 10 points. Learners were given 8 seconds to collect objects, then shown their score for the trial (the net value of all the objects they collected). Each trial consisted of twenty objects, distributed in clusters of five. Players only had enough time to visit a single cluster, which created a two-stage choice: choosing a cluster, then collecting objects from that cluster. Examples of resulting trajectories are shown in Figure \ref{game_ui}(c).  The experiment consisted of instructions and practice, followed by 10 trials.

We paired learners with a teacher, who watched the learner play and was given an opportunity to communicate after each trial. We split teachers into two ``Visibility'' conditions, giving them either ``Full'' or ``Partial'' visibility of the learner's display. ``Partial'' teachers did not see the shapes and colors, giving them less shared context with the learner. We further split teachers into two ``Communication'' conditions, allowing teaching via ``Chat'' or by ``Demonstration.'' This gave us four multiplayer conditions: ``Chat-Full,'' ``Chat-Partial,'' ``Demo-Full'', and ``Demo-Partial.'' This allowed us to study different forms of communication (``Demo'' vs ``Chat'') with more or less shared perception (``Full'' vs ``Partial'' visibility). We hypothesized ``Chat'' pairs would be drastically affected by ``Partial'' visibility, whereas ``Demo'' pairs would not. 

\subsection{Teaching}
\subsubsection{Target Concept}
\begin{figure}[hb!]
\begin{center}
\includegraphics[width=8cm]{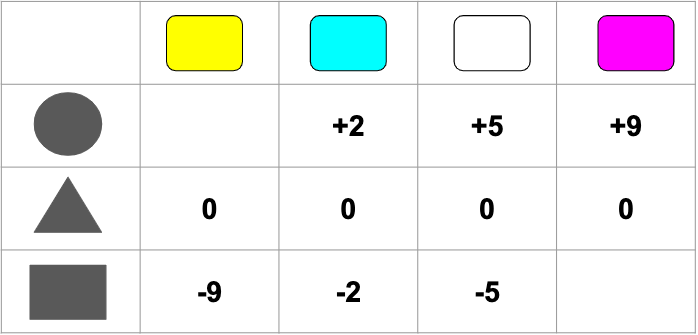}
\end{center}
\caption{The concept to be taught: a mapping of object value to shape and color. Pink and yellow were counterbalanced. Representative values are shown here, but the actual map used a range (i.e. pink circles could be worth 8-10 points). Applying the value mapping to Figure~\ref{game_ui}(b) results in Figure~\ref{game_ui}(a).}  
\label{value_map}
\end{figure}

Participants' scores depended on the learner collecting positive objects and avoiding negative ones. This required them to learn a mapping of shape and color to underlying object values. We designed the mapping to combine independent perceptual dimensions (shape and color) in a non-intuitive manner. Positive objects were rendered as circles, zero-valued objects as triangles, and negative objects as squares. Score was encoded by a spectrum: pink-white-cyan-white-yellow. We counterbalanced the high and low values, presenting half of the players with pink-cyan-yellow and half with yellow-cyan-pink. The full mapping is shown in Figure \ref{value_map} and can be seen in practice in Figure \ref{game_ui}(a) and (b). We subsequently infer learners' beliefs about the concept by estimating their feature-based object value from their choice behavior (see Results: Utility Representation Estimation). 

\subsubsection{Visibility}
Teachers were given ``Partial'' or ``Full'' visibility of the learner’s perceptual state.  “Partial” visibility teachers were shown only the interface in Figure~\ref{game_ui}(b). They could watch the learner play and see the values of all of the objects. However, they did not know what shapes and colors the learner saw, and thus could not teach the concept from the player's percepts. ``Full'' visibility teachers could see both Figure~\ref{game_ui}(a)~and~(b). They could thus infer and teach the value function directly. 
\subsubsection{Communication}
After each round, teachers communicated via demonstrations (``Demo'') or language (``Chat''). ``Demo'' teachers replayed the same level, with the same timing and scoring conditions, while the learner watched. Both parties were informed the teacher's score would be visible to the learner, but would not count. Teachers could thus perform informative actions without repercussion, including choosing negative or zero-valued objects. ``Chat'' teachers were given a chat interface and could send one-way free-form text to the learner. There were no constraints on the amount of time taken, content, length, or number of messages. Teachers were required to send at least one message before advancing.
\subsubsection{Single Player Conditions}
In addition to the multiplayer conditions, we ran two single-player ``Solo'' conditions as controls. ``Solo'' players were divided into two Visibility conditions. ``Solo-Full'' players were shown the object values (Figure~\ref{game_ui}(b)), ``Solo-Partial'' players were shown the learner display (Figure~\ref{game_ui}(a)). ``Solo-Full'' players are used to baseline perfect concept knowledge, whereas ``Solo-Partial'' provide a baseline for incidental learning occurring without any  teaching.

\subsection{Experiment Methods}
\subsubsection{Participants}
We recruited 547 participants via Amazon Mechanical Turk using psiTurk \cite{gureckis_2016}. Participants were paid \$1.50 and a score-based bonus of up to \$1.00. Teachers and learner pairs received the same bonus based on the learner's score. Of the 547 participants, 111 were assigned to a single-player condition and completed all 10 trials. 400 of the remaining participants were matched with a partner, yielding 200 pairs, of whom 167 completed all 10 trials. Dropout ranged from 9\% to 25\% by condition, but the difference was not significant ($\chi^2(3, N=200) = 5.65, p = .13$). We filtered out 3 pairs for non-participation (one learner and one ``Demo'' teacher who never collected an object, and one chat teacher who only sent blank messages). This left 164 pairs and 111 single players, with \textit{n} = 56 for ``Solo-Full''; 55 for ``Solo-Partial''; 47 for ``Demo-Partial''; 36 for ``Demo-Full''; 39 for ``Chat-Partial'', and 42 for ``Chat-Full.''
\subsubsection{Procedure}
Every participant (teachers, learners, and solo players) received instructions and played two practice rounds. They played first with the objects’ values (Figure ~\ref{game_ui}(b)), then with colors and shapes (Figure~\ref{game_ui}(a). This familiarized them with the basic gameplay dynamics. Solo participants proceeded to the experiment. Multiplayer participants were paired, assigned a role, and given condition-specific instructions. They played a practice round with their partner with a simplified value mask, consisting of both the gameplay and teaching phases. Finally, they began the experiment, which consisted of 10 rounds of interleaved gameplay and teaching.

\section{Results}
\begin{figure}[ht!]
\begin{center}
\includegraphics[width=8cm]{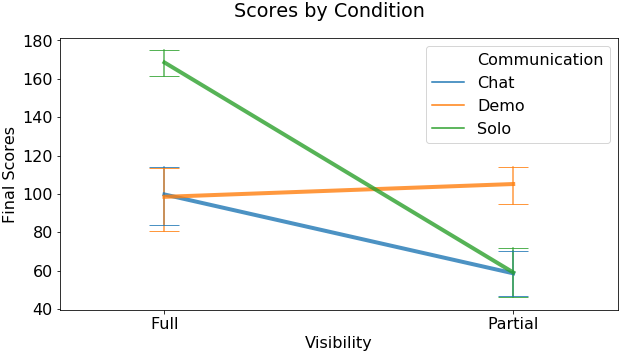}
\end{center}
\caption{Mean scores across the six conditions. Error bars show 95\% CI. Reduction in shared perception (``Full'' to ``Partial'') drastically affected linguistic teaching (``Chat''), rendering it nearly worthless. ``Demo'' teachers performed similarly in both visibility conditions. ``Solo-Full'' provides a baseline for perfect concept knowledge, and ``Solo-Partial'' provides a baseline for no teaching.} 
\label{scores_figure}
\end{figure}

Participants' final scores ranged from -54 to 186 (the highest possible). Scores across  conditions are shown in Figure \ref{scores_figure}. Qualitatively, our six conditions fell into three distinct groups. ``Solo-Full'' players, shown the object value directly as in Figure~\ref{game_ui}(b), fared the best (mean=168, SD=27). We take their actions as representative of perfect concept knowledge. ``Chat-Full'' pairs (mean=100, SD=49), "Demo-Full" pairs (mean=98, SD=52), and "Demo-Partial" pairs (mean=105, SD=35) all performed reasonably well. Learners in these conditions acquired most of the concept over the course of the trials. Finally, ``Chat-Partial'' pairs (mean=59, SD=39) and ``Solo-Partial'' players (mean=59, SD=49) performed the worst. These players struggled to learn the concept; in post-game surveys, many described confusion or incorrect beliefs. 

\FloatBarrier

\subsection{Visibility-Communication Interaction}
Our central hypothesis was the presence of an interaction effect: ``Partial'' visibility of the learner's perceptual state would affect linguistic teaching more than demonstration teaching. To test this hypothesis, we ran a fixed-effects linear regression with contrast coding on the four teaching conditions. The outcome variable was the players' final score, and fixed effects were the teacher's Visibility and Communication, as well as their interaction. All three effects were significant, with the interaction effect being the largest (interaction: $\beta = -47.81\text{, SE = 13.73, }t(160)=3.48,\: p<.01$; communication: $\beta = -22.66,\: SE=6.87,\: t(160)=-3.30,\: p<.01$; visibility: $\beta = -17.22,\: SE=6.87,\: t(160) = -2.51,\: p<.05$). This supports our hypothesis that language is differentially sensitive to misaligned perception: ``Partial'' visibility teachers taught more effectively with demonstrations than with language (see Figure~\ref{scores_figure}). 

Survey results showed a similar interaction effect, with ``Chat-Partial'' learners rating their teachers significantly less helpful. 127 learners filled out a post-experiment survey (\textit{n} = 37 for ``Chat-Full,'' 29 for ``Chat-Partial,'' 25 for ``Demo-Full,'' 36 for ``Demo-Partial''). Learners rated how helpful their teacher was (a six-point scale ranging from ``Very Unhelpful'' to ``Very Helpful''). Mean ratings were 4.83 for ``Chat-Full,'' 3.83 for ``Chat- Partial,'' 4.16 for ``Demo-Full,'' and 4.58 for ``Demo-Partial''; a contrast-coded fixed-effects linear regression on Communication, Visibility, and interaction effects showed significance for their interaction ($\beta = -1.43,\text{ SE=.49, }t(123)=-2.92,\: p<.01$) but not for other effects.


\subsection{Teaching Strategies}
\subsubsection{Demonstration Teaching}
``Demo'' teachers in both conditions adopted similar strategies, with most teachers simply playing optimally (i.e. getting the highest possible score). We assessed difference between the ``Full'' and ``Partial'' teachers with a two-sided T-test on cumulative scores up to the ninth level, which was the last opportunity for teaching. There was not a significant difference between the conditions ($t(82) = -1.61, p=.11$). ``Demo'' teachers resembled ``Solo-Full'' players. A two-sided T-test for ninth-level cumulative scores between the demonstrations and ``Solo-Full'' players did not show a significant difference ($t(137) = -.94, p=.35$). Finally, analysis of the sub-optimal demonstration rounds did not reveal any clear strategies (e.g. collecting a single object to communicate its value). 

\subsubsection{Linguistic Teaching}
``Chat'' teachers in the ``Full'' and ``Partial'' visibility conditions used very different language. After lemmaizing and filtering stop words, the most common tokens and bigrams in the two conditions are shown in Tables 1 and 2. To facilitate comparison, we coded a subset of common words into ``Shapes'' (circle, triangle, square), ``Colors'' (pink, white...), ``Numbers'' (zero, 0, one, 1...), and ``Relational'' (bottom, right, first, second...).

``Chat-Full'' teachers generally taught the concept using the visual features they shared with learners (e.g. ``pink is high positive, turquoise circle is low positive''). They used shape terms (11.1\% of all tokens, versus .2\% for ``Partial'') and color terms (13.0\% of all tokens, versus 1.7\% for ``Partial''). The most common bigram was a reference to the most valuable shape-color combination (because pink and yellow were counterbalanced, this was ``pink''  or ``yellow'' + ``circle''). In contrast, ``Chat-Partial'' teachers used spatial and behavioral references to specific objects, then communicated their value (e.g. ``first one was -7, the others were small amounts 2,1,1'') or provided high-level information (e.g. ``top left corner is the best''). They used relational terms (19.2\% of all tokens, vs 5.3\% for ``Full'') and tended to communicate specific numbers (11.4\% of all tokens, versus 4.6\% for ``Full''). 

\begin{table}[t!]
\begin{center} 
\vskip 0.12in
\begin{tabular}{llll}
\hline
Token   &  Count & Bigram  & Count \\
\hline
circle  &  120 & good job & 44 \\
good & 102 & yellow circle & 27 \\
job &  69 & white circle & 25\\
yellow & 62 & circle worth & 21 \\
white & 68 & pink circle  & 20\\
\hline
\end{tabular}
\label{full_viz_chat}
\caption{Top five tokens and bigrams in the ``Chat-Full'' condition. Teachers used shared visual perception to communicate generalized information. Due to color counterbalancing, ``yellow circle'' and ``pink circle'' both refer to the most valuable color-shape combination for teachers. This feature was correspondingly weighed heavily by ``Chat-Full'' learners (see Figure~\ref{utility_coefficient_results}, right).} 
\end{center}
\end{table}

\begin{table}[t!]
\begin{center} 
\vskip 0.12in
\begin{tabular}{llll}
\hline
Token   &  Count & Bigram  & Count \\
\hline
right  &  92 & bottom right & 26 \\
was & 82 & lower right & 20 \\
good &  74 & great job & 19\\
one & 62 & good job & 18 \\
left & 56 & hand corner  & 15 \\
\hline
\end{tabular}
\label{partial_viz_chat}
\caption{Top five tokens and bigrams in the``Chat-Partial'' condition. Teachers used shared spatial relationships or the player's prior behavior (e.g. ``the first one was good'') to reference specific objects.} 
\end{center} 
\end{table}

The corpora arising from the two conditions reflect the shared context between the two parties. ``Chat-Full'' teachers used visual features to communicate generalized information about relative object values. They almost never referenced specific objects. With less shared context, ``Chat-Partial'' teachers fell back to spatial relationships or the player's prior behavior to ground information.

\subsection{Estimation of Learned Utility Representations}
The previous analyses focused on the effect of different conditions on behavior, but language and demonstration may also affect the \textit{representations} that people learn in systematic ways. Although demonstration is robust, it can also be imprecise (e.g., if a teacher collects a pink circle, the learner faces ambiguity about whether this was due to the object being pink, circular, or both). In contrast, language can be  precise and expressive (e.g. ``one pink circle is worth 2x 1 white circle''). Put another way, language gives teachers more control over the \emph{scope} of what is communicated.

Players' actions demonstrated preferences for different subsets of possible objects. Figure \ref{game_ui}(c) provides an illustration of the trajectories individual players took, each indicating a different belief about which set of objects was most valuable. To assess learned representations, we modeled the rewards that individual learners associate with each feature based on observed choice behavior.

\subsubsection{Model} We modeled participants' behavior using a nested logistic choice model to characterize participants' sequential choices on each trial~\cite{mcfadden1974conditional, Train2003}, similar to procedures for estimating preferences from sequential behavior in machine learning (e.g., inverse reinforcement learning, ~\cite{abbeel_2004}). We treat each trial as a two-stage decision process: participants (1) choose a cluster of objects to move towards, then (2) collect or avoid each object in that cluster. We assume that people prefer to encode a \textit{sparse} representation of the utility function in that they will tend to use only a few of the 19 available shape and color features to choose objects. 

Formally, a particular trial consists of a set of $N$ objects $X_{1:N}$, each of which has $K$ binary features defined by a feature function $\phi: X \rightarrow \{0, 1\}^K$. We used colors (cyan, white, yellow, pink), shapes (square, circle, triangle), and their conjunctions (e.g., pink circle, yellow triangle, etc) as features. Thus, the feature space is size $K = 4+3+4\times3 = 19$. The utility associated with an object $x_i$ is determined by feature-utilities, $\theta$, which linearly combine with an object's features to produce a utility for that object: $U_\theta(x) = \phi(x)^\top \theta$. Given that a participant is in a cluster containing an object $x_i$, the event that they pick up the object ($y_i = 1, y_i \in \{0, 1\}$) has a probability defined by a binary logit model:
\begin{equation}
    p_\theta(y_i = 1 \mid x_i) \propto e^{U_\theta (x_i)}
\end{equation}
Assuming these object choice probabilities, the utility of a cluster of objects $c_j$ is defined by the expected utility of that cluster:
\begin{equation}
    U_\theta(c_j) = \sum_{x_i \in c_j} U_\theta(x_i)p_\theta(y_i = 1 \mid x_i)
\end{equation}
This utility is then used in a multinomial logit model to define cluster-choice probabilities under a utility function, $p_\theta(c_j) = \exp\{U_\theta(c_j)\}$. Thus, given utility weights $\theta$, the probability of a cluster choice $c_j$ and choices $y_{1:N}$ over objects $x_{1:N}$ is:
\begin{equation}
    p(c_j, x_{1:N}, y_{1:N} \mid \theta) = 
    p_\theta(c_j) \prod_{x_i \in c_j} p_\theta(y_i \mid x_i)
\end{equation}

Finally, to model a preference for sparser utility representations, we used a Laplace prior over each utility weight, $p(\theta_k; \lambda) = \frac{\lambda}{2}e^{-\lambda|\theta_k|}$, where $\lambda$ is a scale parameter (in this work we set $\lambda = 1$). This prior corresponds to L1 regularization~\cite{murphy2013} and encourages values of $\theta$ towards 0. We then solved for the \textit{maximum a posteriori} (MAP) estimate of utility weights:
\begin{equation}
    \hat{\theta}_{\text{MAP}} = \argmax_{\theta} p(c_j, x_{1:N}, y_{1:N} \mid \theta)p(\theta)
\end{equation}
We tested several optimizers and found that Powell's conjugate direction method~\cite{1964Powell} produced the most consistent and optimal results, which we report below.

\subsubsection{Modeling Results} 
\begin{figure}
\begin{center}
\includegraphics[width=8cm]{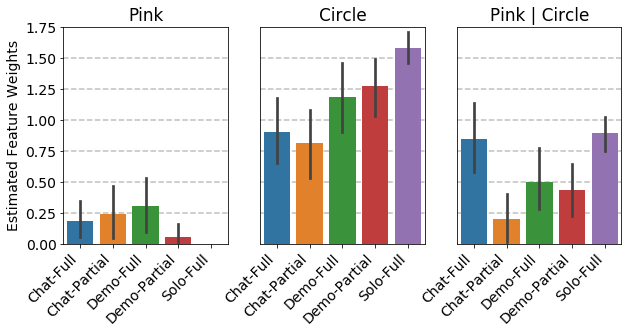}
\caption{Feature coefficients of the utility functions estimated from learner behavior in Levels 8-10. The ``Solo-Full'' condition is our baseline for perfect concept knowledge. Left: learners displayed low preference for ``Pink'' (expected, due to the presence of zero-value pink triangles). Center: learners in the ``Demo'' conditions favored ``Circle`` (all colors) at higher rate, suggesting more robust acquisition of a simpler concept. Right: only learners in the ``Chat-Full'' condition weighed the ``Pink-Circle'' conjunction feature optimally; this color-shape feature was also the most common bigram in the ``Chat-Full'' corpus (see Table 1).} 
\label{utility_coefficient_results}
\end{center}
\end{figure}

In our analysis, we estimated MAP utility functions based on participants' object choices. This allows us to compare utility functions across conditions to infer differential impacts on learning. We model estimated feature coefficients for each participant over the last three trials and plot the resulting weights on ``Pink'', ``Circle'', and ``Pink-Circle'' in Figure~\ref{utility_coefficient_results}. As pink circles were the most valuable objects in the game, these coefficients reflect a critical component of the learner's acquisition of the target concept (Figure~\ref{value_map}). Because ``Solo-Full'' players saw only object values and generally played optimally, their weights reflect perfect knowledge of the concept.

Learners across all conditions demonstrated a weak preference for ``Pink'' alone (Figure~\ref{utility_coefficient_results}, left). This is likely an incorrect generalization of ``Pink'' across circles and triangles, as the presence of zero-value pink triangles should nullify the significance of ``Pink'' as a feature.  In contrast, ``Solo-Full'' participants' collection patterns showed no preference for pink. No learner condition weighed the ``Circle'' feature sufficiently, but the ``Demo'' conditions placed a higher weight than others (Figure~\ref{utility_coefficient_results}, center). A linear regression on the multiplayer conditions with contrast-coded Communication, Visibility, and their interaction as factors showed a significant effect of Communication ($\beta = -0.37,\text{ SE=.14, }t(158)=-2.70,\: p<.01$); other factors were not significant. Finally, only learners in the ``Chat-Full'' condition placed appropriate weight on the more sophisticated ``Pink-Circle'' rule (Figure~\ref{utility_coefficient_results}, right). A linear regression on the multiplayer conditions with contrast-coded Communication, Visibility, and their interaction as factors showed significant effects of both Visibility ($\beta = -0.36,\text{ SE=.12, }t(158)=-2.98,\: p<.01$) and the Communication-Visibility interaction ($\beta = -0.58,\text{ SE=.24, }t(158)=-2.39,\: p<.05$).

Overall, learners in the ``Demo'' conditions appear to have acquired a broader set of features but not the quantitative relationships between them-- in some sense, a more complete but less precise version of the concept. This pattern is supported by the observed teaching behavior. Most ``Demo'' teachers played optimally and thus necessarily collected all of the circles in whichever cluster they chose, which could reasonably cause learners to weight generally accurate features (i.e. ``Circle'') but make it harder to learn critical details (e.g. that the teacher collected three circles and scored 18, but two were worth 8 and one was worth 2). In contrast, ``Full-Chat'' teachers prioritized and communicated pieces of the concept that they felt were most relevant (e.g. in the words of our favorite teacher, ``yellow circle $\vert$ yellow circles $\vert$ yellow circles $\vert$ tellow circles $\vert$ any other circles too''). 

\section{Discussion}
Communication rests on a foundation of shared context. However, different \textit{forms} of communication rely on different contextual information. In our experiment, we ablated a critical piece of visual information, changing the perceptual context shared by teacher and learner. This allowed us to study the interaction between forms of communication and shared context.

Our results highlight a discrepancy between demonstrations and explanation: linguistic communication relies far more on shared percepts. ``Demo'' teachers in both visibility conditions adopted similar strategies, while ``Chat'' teachers used very different language. ``Chat-Full'' teachers referenced shared visual-perceptual attributes to communicate the target concept, whereas ``Chat-Partial'' teachers struggled to establish shared references and gave single object values rather than generalized information. While we induced perceptual misalignment to vary context, a similar dynamic may apply to abstract domains: effectively communicating the Theory of Relativity requires knowing which conceptual building blocks the learner has access to. 

Our results also support the intuition that demonstration provides robust but nonspecific teaching, whereas language-- when successful-- conveys more precise information. ``Demo'' learners more reliably acquired the general contours of object valuation but missed more subtle quantitative relationships, such as the value of different color circles. On the other hand, the precision of language was itself a double-edged sword: many ``Chat-Full'' teachers communicated about the most valuable objects (``pink circles'') immediately, but did not give other useful information (``triangles are worthless'') until later, if at all. 

One interpretation of ``Showing'' versus ``Telling'' is that they place the inferential burden on different parties. Demonstration teaching in the form of optimal action is relatively straightforward for the teacher yet poses challenges for the learner, who must infer the rationale behind the act. In contrast, linguistic teaching is challenging for the teacher, who must infer the gaps in the learner's knowledge and use available shared context to communicate them. 

Finally, the strengths and weaknesses of the teaching methods shown here can inform teaching under disjoint perception. A robustness-precision tradeoff could help explain why demonstration learning has a long history of success in reinforcement learning \cite{abbeel_2004} while language-based teaching remains relatively nascent \cite{narasimhan_2018, reyes_2019}. Our results suggest a hybrid strategy-- starting with demonstrations to illustrate general behavior, followed by specific linguistic corrections-- could take maximum advantage of both channels of communication.  

\section{Acknowledgements}
This work was supported by NSF grant \#1544924, AFOSR grant FA9550-18-1-0077 and grant 61454 from the John Templeton Foundation.

\nocite{ho_2016}
\nocite{shafto_2014} 
\nocite{abbeel_2004} 
\nocite{goodman_2016} 
\nocite{narasimhan_2018} 
\nocite{tomasello} 
\nocite{macGlashan_2017} 
\nocite{rafferty_2016} 
\nocite{reyes_2019} 
\nocite{gureckis_2016} 
\nocite{kiraly_2013} 
\nocite{mcfadden1974conditional} 
\nocite{1964Powell} 
\nocite{goldin_meadow_1999} 
\nocite{gergely_2006} 
\nocite{murphy2013} 
\nocite{Train2003} 
\nocite{Chopra2019}

\bibliographystyle{apacite}

\setlength{\bibleftmargin}{.125in}
\setlength{\bibindent}{-\bibleftmargin}

\bibliography{naive_curriculum}

\end{document}